\documentclass[runningheads]{llncs}
\usepackage[T1]{fontenc}
\usepackage{graphicx}
\usepackage{epstopdf}
\usepackage{subfigure}
\usepackage{cite}
\usepackage{amsmath,amssymb,amsfonts}
\usepackage{algorithmic}
\usepackage{comment}
\usepackage{textcomp}
\usepackage{subfigure}
\usepackage{hyperref}
\usepackage{xcolor}

\usepackage{booktabs} 
\usepackage{array}
\usepackage[utf8]{inputenc}
\usepackage{fontawesome}
\usepackage{graphicx}
\usepackage{float}
\newfloat{figtab}{htb}{fgtb}
\makeatletter
  \newcommand\figcaption{\def\@captype{figure}\caption}
  \newcommand\tabcaption{\def\@captype{table}\caption}
\makeatother
\begin{document}
 \title{Scribble-based 3D Multiple Abdominal Organ Segmentation via Triple-branch Multi-dilated Network with Pixel- and Class-wise Consistency}
 \titlerunning{Scribble-based 3D Multiple Abdominal Organ Segmentation}
 \author{
Meng Han\inst{1} \and
Xiangde Luo\inst{1} \and Wenjun Liao\inst{2} \and Shichuan Zhang\inst{2} \and Shaoting~Zhang\inst{1,3}\and Guotai Wang\inst{1,3,*}
}

\authorrunning{M. Han et al.}
\institute{School of Mechanical and Electrical Engineering, University of Electronic Science and Technology of China, Chengdu, China 
\and Department of Radiation Oncology, Sichuan Cancer Hospital $\&$ Institute, University of Electronic Science and Technology of China, Chengdu, China
\and Shanghai Artificial Intelligence Laboratory, Shanghai, China \\
* Corresponding author:
\email{guotai.wang@uestc.edu.cn}
}
%%%%%%%%%%%%%%%%%%%%%%%%%%%%%%%%%%%%%%%%%%%%
\maketitle              
\begin{abstract}
Multi-organ segmentation in abdominal Computed Tomography (CT) images is of great importance for diagnosis of abdominal lesions and subsequent treatment planning. Though deep learning based methods have attained high performance, they rely heavily on large-scale pixel-level annotations that are time-consuming and labor-intensive to obtain. 
Due to its low dependency on annotation, weakly supervised segmentation has attracted great attention. 
However, there is still a large performance gap between current weakly-supervised methods and fully supervised learning, leaving room for exploration.
In this work, we propose a novel 3D framework with two consistency constraints for scribble-supervised multiple abdominal organ segmentation from CT. 
Specifically, we employ a Triple-branch multi-Dilated network (TDNet) with one encoder and three decoders using different dilation rates to capture features from different receptive fields that are complementary to each other to generate high-quality soft pseudo labels.  
For more stable unsupervised learning, we use voxel-wise uncertainty to rectify the soft pseudo labels and then supervise the outputs of each decoder. To further regularize the network, class relationship information is exploited by encouraging the generated class affinity matrices to be consistent across different decoders under multi-view projection. Experiments on the public WORD dataset show that our method outperforms five existing scribble-supervised methods.

\end{abstract}
\section{Introduction}
Abdominal organ segmentation from medical images is an essential work in clinical diagnosis and treatment planning of abdominal lesions\cite{luo2022word}.
Recently, deep learning methods based on  Convolution Neural Network (CNN) have achieved impressive performance in medical image segmentation tasks\cite{chen2021deep, wang2019abdominal}. 
However, their success relies heavily on large-scale high-quality pixel-level annotations that are too expensive and time-consuming to obtain, especially for multiple organs in 3D volumes.
Weakly supervised learning with a potential to reduce annotation costs has attracted great attention.
Commonly-used weak annotations include dots\cite{en2022annotation, liang2022tree}, scribbles\cite{liu2022weakly, chen2022scribble2d5,luo2022scribble,liang2022tree}, bounding boxes\cite{dai2015boxsup}, and image-level tags\cite{wei2018revisiting, ru2022learning}. 
Compared with the other weak annotations, scribbles can provide more location information about the segmentation targets, especially for objects with irregular shapes\cite{chen2022scribble2d5}. Therefore, this work focuses on exploring high-performance models for multiple abdominal organ segmentation based on scribble annotations.

Training CNNs for segmentation with scribble annotations has been increasingly studied recently. 
Existing methods are mainly based on pseudo label learning\cite{luo2022scribble, liang2022tree}, regularized losses\cite{kim2019mumford, tang2018regularized, obukhov2019gated} and consistency learning\cite{liu2022weakly, zhang2022cyclemix, gao2022segmentation}.
Pseudo label learning methods deal with unannotated pixels by generating fake semantic labels for learning. 
For example, Luo et al.\cite{luo2022scribble} introduced a network with two slightly different decoders that generate dynamically  mixed pseudo labels for supervision. 
Liang et al.\cite{liang2022tree} proposed to leverage minimum spanning trees to generate low-level and high-level affinity matrices based on color information and semantic features to refine the pseudo labels. 
Arguing that the pseudo label learning may be unreliable, Tang et al.\cite{tang2018regularized} introduced the Conditional Random Field (CRF) regularization loss for image segmentation directly.
Obukhov et al.\cite{obukhov2019gated} proposed to incorporate the gating function with CRF loss considering the directionality of unsupervised information propagation.
Recently, consistency strategies that encourage consistent outputs of the network for the same input under different perturbations have achieved increasing attentions.
Liu et al.\cite{liu2022weakly} introduced transformation-consistency based on an uncertainty-aware mean teacher\cite{cui2019semi} model. 
Zhang et al.\cite{zhang2022cyclemix} proposed a framework 
composed of mix augmentation and cycle consistency.
Although these scribble-supervised methods have achieved promising results, their performance is still much lower than that of fully-supervised training, leaving room for improvement. 

Differently from most existing weakly supervised methods that are designed for 2D slice segmentation with a single or few organs, we propose a highly optimized 3D triple-branch network with one encoder and three different decoders, named TDNet, to learn from scribble annotations for segmentation of multiple abdominal organs. 
Particularly, the decoders are assigned with different dilation rates\cite{wei2018revisiting} to learn features from different receptive fields that are complementary to each other for segmentation, which also improves the robustness of dealing with organs at different scales as well as the feature learning ability of the shared encoder.
Considering the features at different scales learned in these decoders, we fuse these multi-dilated predictions to obtain more accurate soft pseudo labels rather than hard labels\cite{luo2022scribble} that tend to be over-confidence predictions.
For more stable unsupervised learning, we use voxel-wise uncertainty to rectify the soft pseudo labels and then impose consistency constraints on the output of each branch. 
In addition, we extend the consistency to the class-related information level\cite{tung2019similarity} to constrain inter-class affinity for better distinguishing them. Specifically, we generate the class affinity matrices in different decoders and encourage them to be consistent after projection in different views.
% Afterward, we combine the scribbles supervision and two consistency supervision to train the segmentation network end-to-end. 

The contributions of this paper are summarized as follows: 1) We propose a novel 3D Triple-branch multi-Dilated network called TDNet for scribble-supervised segmentation. By equipping with varying dilation rates, the network can better leverage multi-scale context for dealing with organs at different scales. 2) We propose two novel consistency loss functions, i.e., Uncertainty-weighted Soft Pseudo label Consistency (USPC) loss and Multi-view Projection-based Class-similarity Consistency (MPCC) loss, to regularize the prediction from the pixel-wise and class-wise perspectives respectively, which helps the segmentation network obtain reliable predictions on unannotated pixels. 3) Experiments results show our proposed method outperforms five existing scribble-supervised methods on the public dataset WORD\cite{luo2022word} for multiple abdominal organ segmentation.
%%%%%%%%%%%%%%%%%%%
\begin{figure}[t]
\includegraphics[width=\textwidth]{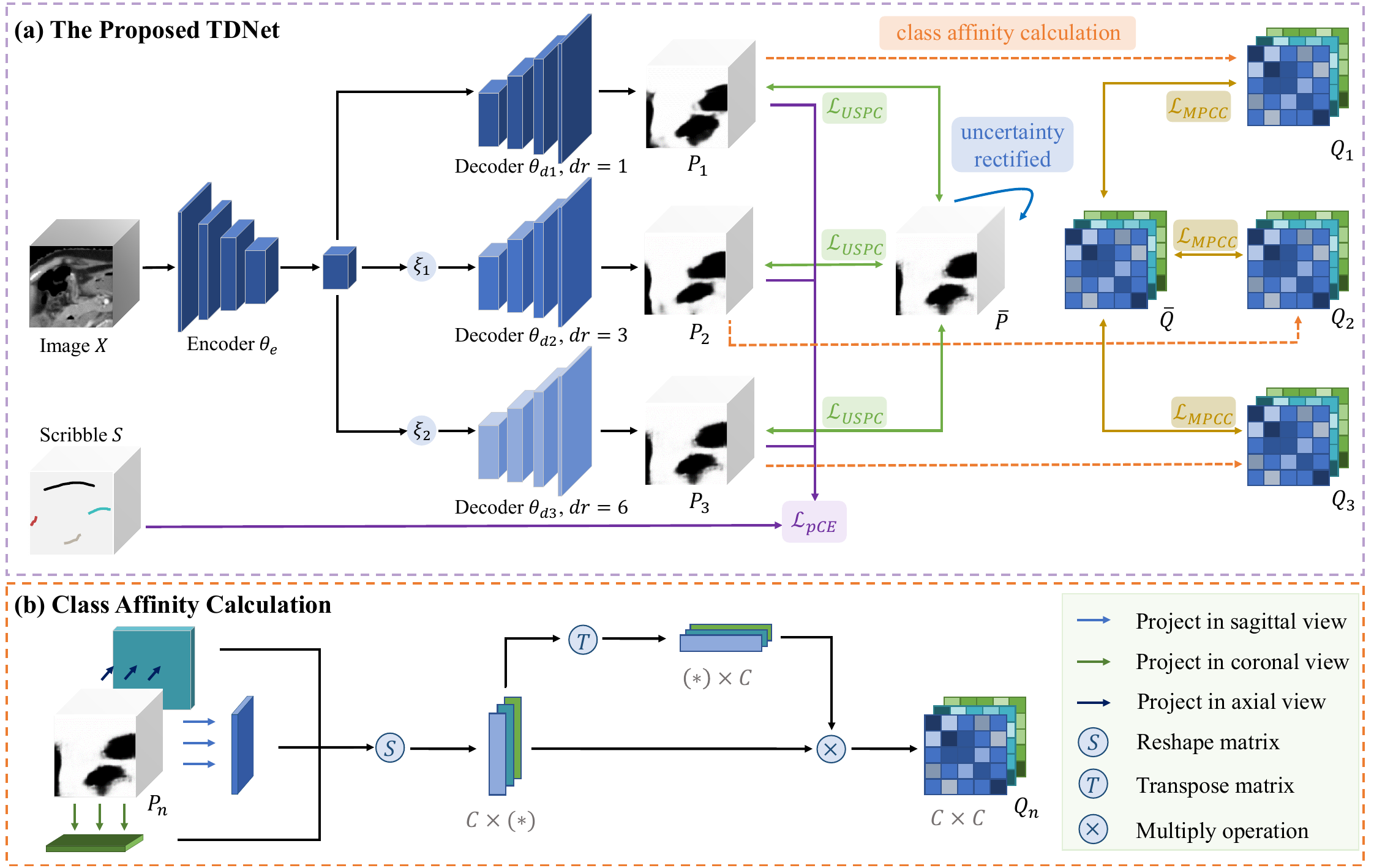}
\vspace{-5mm}
\caption{Overview of the proposed Triple-branch multi-Dilated Network (TDNet) that uses different dilation rates at three decoders. 
The TDNet is optimized by Uncertainty-weighted Soft Pseudo label Consistency (USPC) using the mixed soft pseudo labels and Multi-view Projection-based Class-similarity Consistency (MPCC).  The class affinity calculation process is shown in (b). Best viewed in color.} 
\label{framework}
\end{figure}
%%%%%%%%%%%%%%%%%%%
\section{Method}
Fig. \ref{framework} shows the the proposed framework for scribble-supervised medical image segmentation. We introduce a network with one encoder and three decoders with different dilation rates to learn multi-scale features. 
The decoders' outputs are averaged to generate a soft pseudo label that is rectified by uncertainty and then used to supervise each branch.  
To better deal with multi-class segmentation, a class similarity consistency loss is also used for regularization.  

For the convenience of following description, we first define several mathematical symbols. Let $X, S$ be a training image and the corresponding scribble annotation, respectively. Let $C$ denote the number of classes for segmentation, % $P_1$, $P_2$, $P_3$ be the outputs from decoders. 
and $\Omega = \Omega_{S} \cup \Omega_{U}$ denote the whole set of voxels in $X$, where $\Omega_{S}$ is the set of labeled pixels annotated in $S$, and  $\Omega_{U}$ is the unlabeled pixel set. 
\subsection{Triple-branch Multi-dilated Network (TDNet)}\label{sec:tdnet}
As shown in Fig.~\ref{framework}(a), the proposed TDNet consists of a shared encoder ($\theta_e$) and three independent decoders ($\theta_{d1},\theta_{d2}, \theta_{d3}$) with different dilation rates to mine unsupervised context from different receptive fields.
Specifically, decoders using convolution with small dilation rates can extract detailed local features but their receptive fields are small for understanding a global context. 
Decoders using convolution with large dilation rates can better leverage the global information but may lose some details for accurate segmentation. 
In this work, our TDNet is implemented by introducing two auxiliary decoders into a 3D UNet\cite{cciccek20163d}. The dilation rate in the primary decoder and the two auxiliary decoders are 1, 3 and 6 respectively, with the other structure parameters (e.g., kernel size, channel number etc.) being the same in the three decoders.
To further introduce perturbations for obtaining diverse outputs, the three branches are initialized with Kaiming initialization, Xavier and Normal initialization methods, respectively. In addition, the bottleneck's output features are randomly dropped out before sending into the auxiliary decoders. The probability prediction maps obtained by the three decoders are denoted as $P_1$, $P_2$ and $P_3$, respectively.

\subsection{Pixel-wise and Class-wise Consistency}
\subsubsection{Uncertainty-weighted Soft Pseudo label Consistency (USPC)}
As the three decoders capture features at different scales that are complementary to each other, an ensemble of them would be more robust than a single branch. Therefore, we take an average of $P_1, P_2, P_3$ to get a better soft pseudo label $\bar{P} = (P_1 + P_2 + P_3)/3$ that is used to supervise each branch during training. 
However, $\bar{P}$ may also contain noises and be inaccurate, and it is important to highlight reliable pseudo labels while suppressing unreliable ones. 
Thus, we propose a regularization term named 
Uncertainty-weighted Soft Pseudo label Consistency (USPC) between $P_n$ ($n$= 1, 2, 3)  and $\bar{P}$:
\begin{equation}
\mathcal{L}_{USPC}=\frac{1}{3} \sum_{n={1,2,3}} \frac{\sum_i w_i KL(P_{n,i} \| \bar{P}_i)}{\sum_i w_i}
\end{equation}
where $\bar{P}_i$ refers to the prediction probability at voxel $i$ in $\bar{P}$, and $\bar{P}_{n,i}$ is the corresponding prediction probability at voxel $i$ in  $\bar{P}_{n}$. $KL$() is the Kullback–Leibler divergence.
$w_i$ is the voxel-wise weight  based on uncertainty estimation: 
\begin{equation}
w_{i} = e^{ \sum_c \bar{P}^c_i log (\bar{P}^c_i)}
\end{equation}
where the uncertainty is estimated by entropy. $c$ is the class index, and $\bar{P}^c_i$ means the probability for class $c$ at voxel $i$ in the pseudo label. Note that a higher uncertainty leads to a lower weight. With the uncertainty-based  weighting, the model will be less affected by unreliable pseudo labels.
\subsubsection{Multi-view Projection-based Class-similarity Consistency (MPCC)}
For multi-class segmentation tasks, it is important to learn inter-class relationship for better distinguishing them.
In addition to using $\mathcal{L}_{USPC}$ for pixel-wise supervision, we consider making consistency on class relationship across the outputs of the decoders as illustrated in Fig. \ref{framework}. 
In order to save computing resources, we project the soft pseudo labels along each dimension and then calculate the affinity matrices, which also strengthens the class relationship information learning.
We first project the soft prediction map of the $n$-th decoder $P_n \in \mathbb{R}^{C \times D \times H \times W} $ in axial view to a tensor with the shape of ${C\times 1\times H\times W}$. It is reshaped into $C\times (WH)$ and multiplied by its  transposed version, leading to a  class affinity matrix $Q'^{axial}_n\in \mathbb{R}^{C \times C}$. A normalized version of $Q'^{axial}_n$ is denoted as $Q^{axial}_n = Q'^{axial}_n/ ||Q'^{axial}_n||$.  Similarly, $P_n$ is projected in the sagittal and coronal views, respectively, and the corresponding normalized class affinity matrices are denoted as $Q^{sagittal}_n$ and $Q^{coronal}_n$, respectively. Here, the affinity matrices represents the relationship between any pair of classes along the dimensions. 
Then we constraint the consistency among the corresponding affinity matrices by Multi-view Projection-based Class-similarity Consistency (MPCC) loss:
\begin{equation}
\mathcal{L}_{MPCC}= \frac{1}{3\times3} \sum_v \sum_{n={1,2,3}} KL(Q^v_n \| \bar{Q}^v)  
\end{equation}
where $v \in $ \{axial, sagittal, coronal\} is the view index, and $\bar{Q}^v$ is the average class affinity matrix in a certain view obtained by the three decoders.
\subsection{Overall Loss Function}
To learn from the scribbles, the partially Cross-Entropy (pCE) loss is used to train the network, where the labeled pixels are considered to calculate the gradient and the other pixels are ignored\cite{tang2018normalized}:
\begin{equation}
\mathcal{L}_{sup}=-\frac{1}{3 \left| \Omega_S \right|}  \sum_{n={1,2,3}} \sum_{i \in \Omega_S} \sum_{c} S^c_{i} \log P^c_{n, i}
\end{equation}
where $S$ represents the one-hot scribble annotation, and $\Omega_S$ is the set of labeled pixels in $S$. 
The total object function is summarized as:
\begin{equation}
    \mathcal{L}_{total} = \mathcal{L}_{sup} +\alpha_t\mathcal{L}_{USPC} + \beta_t \mathcal{L}_{MPCC}
    \label{totalloss}
\end{equation}
where $\alpha_t$ and $\beta_t$ are the weights for the unsupervised losses. Following \cite{liu2022weakly}, we define $\alpha_t$ based on a ramp-up function: $\alpha_t= \alpha \cdot e^{\left(-5\left(1-t / t_{\max }\right)^2\right)}$, where $t$ denotes the current training step and
$t_{max}$ is the maximum training step. We define $\beta_t= \beta \cdot e^{\left(-5\left(1-t / t_{\max }\right)^2\right)}$ in a similar way. In this way, the model can learn accurate information from scribble annotations, which also avoids getting stuck in a degenerate solution due to low-quality pseudo labels at an early stage.

\section{Experiments and Results}
\subsection{Dataset and Implementation Details}
We used the publicly available abdomen CT dataset WORD\cite{luo2022word} for experiments, which consists of 150 abdominal CT volumes from patients with rectal cancer, prostate cancer or cervical cancer before radiotherapy. Each CT volume contains 159-330 slices of $512 \times 512$ pixels, with an in-plane resolution of 0.976 $\times$ 0.976~mm and slice spacing of 2.5-3.0~mm. We aimed to segment seven organs: the liver, spleen, left kidney, right kidney, stomach, gallbladder and pancreas. Following the default settings in \cite{luo2022word}, the dataset was split into 100 for training, 20 for validation and 30 for testing, respectively, where the scribble annotations for foreground organs and background in the axial view of the training volumes had been provided and were used in model training. For pre-processing, we cut off the Hounsfield Unit (HU) values with a fixed window/level of 400/50 to focus on the abdominal organs, and normalized it to $[0, 1]$. We used the commonly-adopted Dice Similarity Coefficient (DSC), 95$\%$ Hausdorff Distance(HD$_{95}$) and the Average Surface Distance (ASD) for quantitative evaluation.

Our framework was implemented in PyTorch\cite{paszke2019pytorch} on an NVIDIA 2080Ti with 11GB memory.
We employed the 3D UNet\cite{cciccek20163d} as the backbone network for all experiments, and extended it with three decoders by embedding two auxiliary decoders with different dilation rates, as detailed in Section~\ref{sec:tdnet}. To introduce perturbations, different initializations were applied to each decoder, and random perturbations (ratio $=(0, 0.5)$) were introduced in the bottleneck before the auxiliary decoders.
The Stochastic Gradient Descent (SGD) optimizer with momentum of 0.9 and weight decay of $10^{-4}$ was used to minimize the overall loss function formulated in Eq.\ref{totalloss}, where $\alpha$=10.0 and $\beta$=1.0  based on the best performance on the validation set.
The poly learning rate strategy\cite{luo2021efficient} was used to decay learning rate online. The batch size, patch size and maximum iterations $t_{\max}$ were set to $1$, $[80,96,96]$ and $6 \times 10^{4}$ respectively.
The final segmentation results were obtained by using a sliding window strategy.
For a fair comparison, we used the primary decoder’s outputs as the final results during the inference stage and did not use any post-processing methods. Note that all experiments were conducted in the same experimental setting. The existing methods are implemented with the help of open source codebase from\cite{wsl4mis2020}.

\begin{table}[t!]
  \centering
  \caption{Quantitative comparison between our method and existing weakly supervised methods on WORD testing set.
  %Mean and standard deviation (subscript) of 3D DSC, ASSD and HD$_{95}$ are reported. 
  $*$ denotes  p-value < 0.05 (paired t-test) when comparing with the second place method \cite{luo2022scribble}.
  The best values are highlighted in bold.}
  % \vspace{-2mm}
  \resizebox{1.0\columnwidth}{!}{
    \begin{tabular}{ccccccccc}
    \toprule
    Organ & FullySup\cite{cciccek20163d}    & pCE   &  TV\cite{javanmardi2016unsupervised}    & USTM\cite{liu2022weakly}  & EM\cite{grandvalet2004semi}    &DMPLS\cite{luo2022scribble} & Ours \\
    \midrule
    Liver & 96.37$_{\pm{0.74}}$ & 86.53$_{\pm{3.61}}$ &  87.22$_{\pm{3.07}}$ & 80.57$_{\pm{3.87}}$ & 89.60$_{\pm{1.72}}$ &   88.89$_{\pm{2.54}}$ & \textbf{93.31}$_{\pm{\textbf{0.85}}}$$^*$ \\
    spleen & 95.42$_{\pm{1.55}}$ & 86.81$_{\pm{5.75}}$ &  82.95$_{\pm{7.30}}$ & 86.49$_{\pm{4.71}}$ & 87.76$_{\pm{5.65}}$ & 89.19$_{\pm{3.93}}$ &\textbf{91.77}$_{\pm{\textbf{3.19}}}$$^*$ \\
    kidney(L) & 94.95$_{\pm{1.58}}$ & 86.25$_{\pm{4.46}}$  &  83.78$_{\pm{4.60}}$ & 87.15$_{\pm{4.21}}$ & 87.29$_{\pm{3.88}}$ &90.14$_{\pm{2.98}}$& \textbf{92.34}$_{\pm{\textbf{2.30}}}$$^*$\\
    kidney(R) & 95.33$_{\pm{1.34}}$ & 89.41$_{\pm{2.97}}$ & 89.38$_{\pm{3.15}}$ & 89.26$_{\pm{2.35}}$ & 81.32$_{\pm{4.79}}$ & 90.93$_{\pm{2.15}}$&\textbf{92.54}$_{\pm{\textbf{1.95}}}$$^*$ \\
    stomach & 90.08$_{\pm{4.42}}$ & 61.09$_{\pm{12.20}}$ & 62.64$_{\pm{12.74}}$ & 77.33$_{\pm{6.19}}$ & 77.74$_{\pm{7.34}}$ &77.06$_{\pm{7.39}}$& \textbf{85.82}$_{\pm{\textbf{4.19}}}$$^*$\\
    gallbladder & 75.33$_{\pm{13.21}}$ & 56.61$_{\pm{20.12}}$ &   44.06$_{\pm{19.92}}$ & 63.94$_{\pm{17.17}}$ & 65.83$_{\pm{16.80}}$ &\textbf{70.25}$_{\pm{\textbf{15.57}}}$& 69.01$_{\pm{16.53}}$ \\
    pancreas & 80.90$_{\pm{7.67}}$ & 61.55$_{\pm{10.45}}$ & 65.31$_{\pm{9.20}}$ & 67.24$_{\pm{9.49}}$ & 73.52$_{\pm{7.54}}$ &75.02$_{\pm{7.53}}$& \textbf{75.40}$_{\pm{\textbf{7.51}}}$ \\
    \midrule
    avg DSC$(\%)$ & 89.77$_{\pm{4.36}}$ & 75.46$_{\pm{8.51}}$ &  73.62$_{\pm{8.57}}$ & 78.85$_{\pm{6.86}}$ & 80.44$_{\pm{6.82}}$ &83.07$_{\pm{6.01}}$& \textbf{85.74}$_{\pm{\textbf{5.22}}}$$^*$ \\
    avg ASD(mm) & 1.60$_{\pm{1.56}}$ & 25.11$_{\pm{11.59}}$ &  31.01$_{\pm{12.41}}$ & 18.24$_{\pm{9.29}}$ & 16.17$_{\pm{8.35}}$ & 7.77$_{\pm{6.33}}$&\textbf{2.33}$_{\pm{\textbf{1.76}}}$$^*$ \\
    avg HD$_{95}$(mm) & 5.71$_{\pm{5.36}}$ & 77.56$_{\pm{36.80}}$ &  98.61$_{\pm{39.93}}$ & 61.43$_{\pm{34.79}}$ & 50.90$_{\pm{29.92}}$ & 24.00$_{\pm{20.08}}$&\textbf{7.84}$_{\pm{\textbf{5.84}}}$$^*$ \\
    \bottomrule
    % \vspace{-6mm}
    \end{tabular}
    }
  \label{quantative comparision with sota}%
\end{table}%
\begin{figure}[t!]
\includegraphics[width=\textwidth]{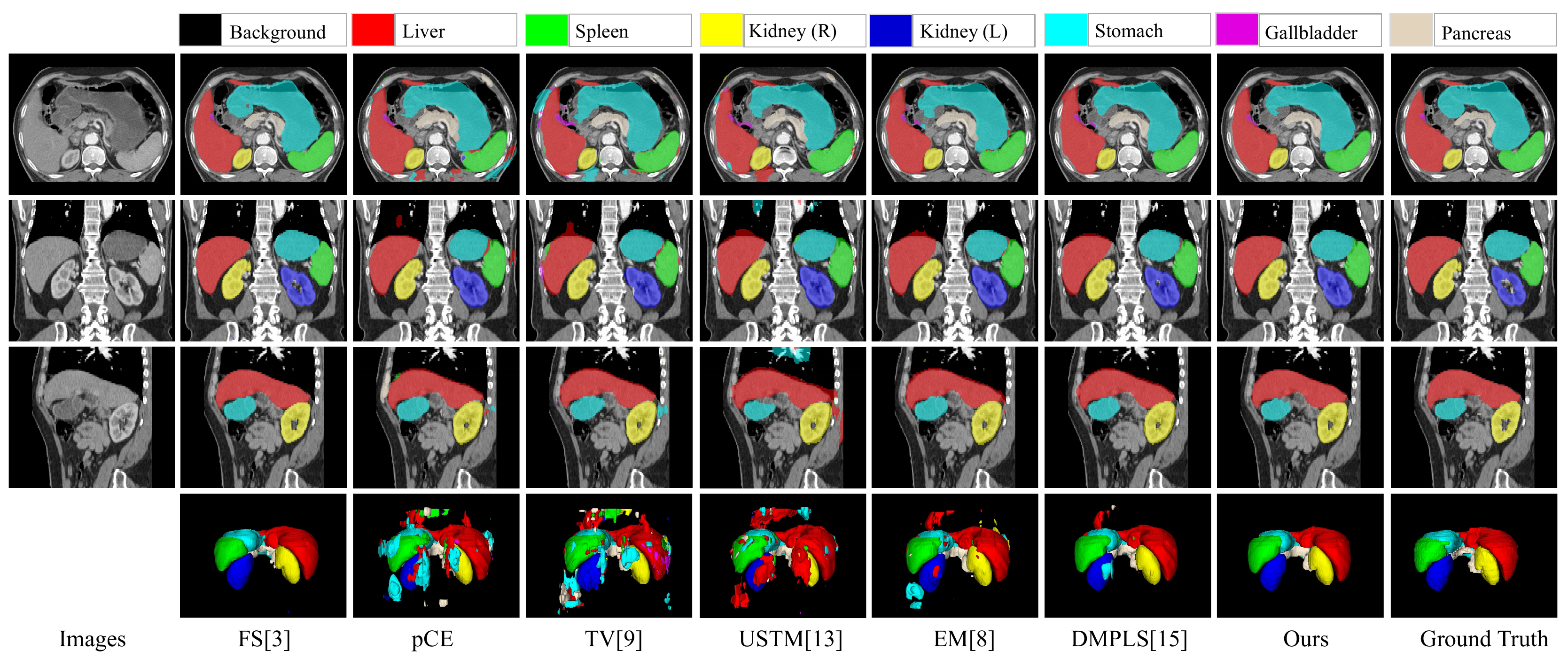}
\vspace{-7mm}
\caption{Visual comparison between  our method and other weakly supervised methods. Best view in color.} 
\label{visual_results}
\end{figure}
\subsection{Comparison with other methods}
We compared our method with five weakly supervised segmentation methods with the same set of scribbles, including pCE only~\cite{lin2016scribblesup}, Total Variation Loss (TV)~\cite{javanmardi2016unsupervised}, Uncertainty-aware Self-ensembling and Transformation-consistent Model (USTM)~\cite{liu2022weakly}, Entropy Minimization (EM)\cite{grandvalet2004semi} and Dynamically Mixed Pseudo Labels Supervision (DMPLS)\cite{luo2022scribble}. 
They were also compared with the upper bound by using dense annotation to train models (FullySup)\cite{cciccek20163d}. 
% All these methods were implemented by 3D Unet\cite{cciccek20163d} as the backbone. 
The results in Table~\ref{quantative comparision with sota} show that our method leads to the best DSC, ASD and HD$_{95}$. 
%(more results about ASD and HD$_{95}$ are showed in supplementary materials). 
Compared with the second best method DMPLS~\cite{luo2022scribble}, the average DSC was increased by 2.67 percent points, and the average ASD and HD$_{95}$ were decreased by 5.44 mm and 16.16 mm, respectively. 
It can be observed that TV\cite{javanmardi2016unsupervised} obtained a worse performance than pCE, which is mainly because that method classifies pixels by minimizing the intra-class intensity variance, making it difficult to achieve good segmentation  due to the low contrast. 
Fig.~\ref{visual_results} shows a visual comparison between our method and the other weakly supervised methods on the WORD dataset (word$\_$0014.nii). It can be obviously seen that the results obtained by our method are closer to the ground truth, with less mis-segmentation in both slice level and volume level.
%%%%%%%%%%%%%%%%%%%%%%%%%%%%
\begin{figtab}[t!]
\begin{minipage}[htbp]{.55\linewidth}
  \centering
  \tabcaption{Ablation study of our proposed method on WORD validation set. N$(s)$ and N$(d)$ means N decoders with the same and different dilation rates, respectively. $\mathcal{L}_{sup}$ is used by default. The best values are highlighted in bold.
  }
  \vspace{2mm}
   \resizebox{1.0\columnwidth}{!}{
    \begin{tabular}{ccccc}
    \toprule
    Decoder  & Loss  & DSC($\%$)  &  ASD(mm)  & HD$_{95}$(mm) \\
    \midrule
    % \vspace{1mm}
    $1(s)$     &       &     74.70$_{\pm{8.68}}$  &    25.51$_{\pm{10.12}}$   &  79.98$_{\pm{30.39}}$  \\ %Val
    % \vspace{1mm}
    % $3(s)$     &       &     78.02$_{\pm{8.39}}$  &    20.99$_{\pm{10.81}}$   &  72.76$_{\pm{42.64}}$  \\ %val
    $3(s)$  &   $\mathcal{L}_{USPC} (-\omega)$    &  81.92$_{\pm{8.04}}$     &   9.40$_{\pm{6.79}}$     & 31.11$_{\pm{20.92}}$  \\
    %\vspace{1mm}
    $3(d)$  &   $\mathcal{L}_{USPC} (-\omega)$    &   82.57$_{\pm{7.28}}$     &  3.34$_{\pm{2.67}}$      &  9.26$_{\pm{7.26}}$ \\
    % \vspace{1mm}
    $3(d)$  &    $\mathcal{L}_{USPC}$   &   84.21$_{\pm{6.99}}$    & 2.82$_{\pm{2.71}}$      & 8.25$_{\pm{6.36}}$ \\
    % \vspace{1mm}
    $3(d)$  &  $\mathcal{L}_{USPC} + \mathcal{L}_{MPCC}$  &  \textbf{84.75}$_{\pm{\textbf{7.01}}}$     &  \textbf{2.64}$_{\pm{\textbf{2.46}}}$     & \textbf{7.91}$_{\pm{\textbf{5.93}}}$  \\
    % \vspace{1mm}
    $2(d)$  &     $\mathcal{L}_{USPC} + \mathcal{L}_{MPCC}$  &   84.18$_{\pm{6.84}}$    &   2.85$_{\pm{2.38}}$    & 8.56$_{\pm{6.36}}$ \\
    $4(d)$  &     $\mathcal{L}_{USPC} + \mathcal{L}_{MPCC}$  &  83.51$_{\pm{7.01}}$     &   2.88$_{\pm{2.49}}$     & 8.58$_{\pm{6.53}}$ \\
    \bottomrule
    \end{tabular}%
    \label{ablation_study}%
    }
\end{minipage}\quad
%%%%%%%%%%%%%%%%%%%%%
\begin{minipage}[htbp!]{.44\linewidth}
\centering
\includegraphics[width = 0.8\linewidth]{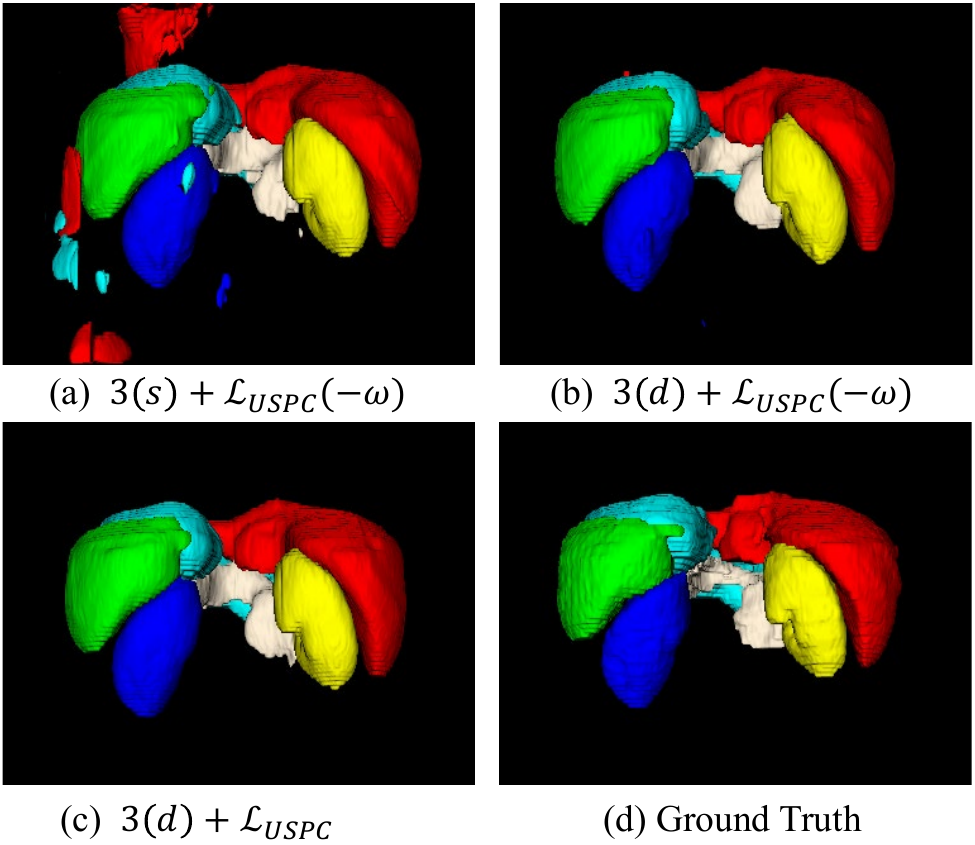}
\vspace{-3mm}
\figcaption{Visualization of the improvement obtained by using different dilation rates and uncertainty rectifying. Best viewed in color.} 
\label{ablation_fig}
\end{minipage}
%%%%%%%%%%%%%%
\end{figtab}
%%%%%%%%%%%%%%%%%%%%%%%
\subsection{Ablation experiment}
We then performed ablation experiments to investigate the contribution of each part of our method, and the quantitative results on the validation set are shown in Table~\ref{ablation_study}, where 
$\mathcal{L}_{USPC} (-\omega)$ means using $\mathcal{L}_{USPC}$ without pixel-wise uncertainty rectifying. Baseline refers to a triple-branch model with different initializations and random feature-level dropout in the bottleneck, supervised by pCE only.
It can be observed that by using $\mathcal{L}_{USPC}(-\omega)$ with mutiple decoders, the model segmentation performance is greatly enhanced with average DSC increasing by 7.70\%, ASD and HD$_{95}$ decreasing by 16.11~mm and 48.87~mm, respectively.
By equipping each decoders with different dilation rates, the model's performance is further improved, especially in terms of ASD and HD$_{95}$, which proves our hypothesis that learning features from different scales can improve the segmentation accuracy. 
Replacing $\mathcal{L}_{USPC}(-\omega)$ with $\mathcal{L}_{USPC}$ further improved the DSC to 84.21\%, and reduced the ASD and HD$_{95}$ by 0.52~mm and 1.01~mm through utilizing the uncertainty information. 
Visual comparison in Fig.~\ref{ablation_fig} demonstrates that over-segmentation can be mitigated by using different dilation rates in the three decoders, and using the uncertainty-weighted pseudo labels can further improve the segmentation accuracy with small false positive regions removing. 

Additionally, Table~\ref{ablation_study} shows that combining $\mathcal{L}_{USPC}$ and $\mathcal{L}_{MPCC}$ obtained the best performance, where the average DSC, ASD and HD$_{95}$ were 84.75\%, 2.64~mm and 7.91~mm, respectively,  which demonstrates the effectiveness of the proposed class similarity consistency. 
In order to find the optimal number of decoders, we set the decoder number to  2, 3 and 4 respectively. The quantitative results in the last three rows of Table~\ref{ablation_study} show that using three decoders outperformed using two  and four decoders.

\section{Conclusion}
In this paper, we proposed a scribble-supervised multiple abdominal organ segmentation method consisting of a 3D triple-branch multi-dilated network with two-level consistency constraints. By equipping each decoder with different dilation rates, the model leverages features at different scales to obtain high-quality soft pseudo labels. 
In addition to mine knowledge from unannotated pixels, we also proposed USPC Loss and MPCC Loss to learn unsupervised information from the uncertainty-rectified soft pseudo labels and class affinity matrix information respectively. 
Experiments on a public abdominal CT dataset WORD demonstrated the effectiveness of the proposed method, which outperforms five existing scribble-based methods and narrows the performance gap between weakly-supervised and fully-supervised segmentation methods.
In the future, we will explore the effect of our method on sparser labels, such as a volumetric data with scribble annotations on one or few slices.

\subsubsection{Acknowledgements}
This work was supported by the National Natural Science Foundation of China (No.62271115), Science and Technology Department of Sichuan Province, China (2022YFSY0055) and Radiation Oncology Key Laboratory of Sichuan Province Open Fund (2022ROKF04).

\bibliographystyle{splncs04}
\bibliography{mybib}

\end{document}